\documentclass[10pt,twocolumn,letterpaper]{article}

\usepackage{iccv}
\usepackage{times}
\usepackage{epsfig}
\usepackage{graphicx}
\usepackage{amsmath}
\usepackage{amssymb}
\usepackage{multirow}

\usepackage[breaklinks=true,bookmarks=false]{hyperref}

\iccvfinalcopy 

\def\iccvPaperID{****} 

\ificcvfinal\fi

\begin{document}

\title{From Two to One: A New Scene Text Recognizer with \\Visual Language Modeling Network}

\author{Yuxin Wang$^1$, Hongtao Xie$^1$, Shancheng Fang$^1$\thanks{Corresponding author} , Jing Wang$^2$, Shenggao Zhu$^2$ and Yongdong Zhang$^1$\\
$^1$University of Science and Technology of China\\
$^2$Huawei Cloud \& AI\\
{\tt\small wangyx58@mail.ustc.edu.cn, \{htxie,fangsc,zhyd73\}@ustc.edu.cn}\\
{\tt\small \{wangjing105,zhushenggao\}@huawei.com}
}
\maketitle
\ificcvfinal\thispagestyle{empty}\fi

\begin{abstract}
    In this paper, we abandon the dominant complex language model and rethink the linguistic learning process in the scene text recognition. Different from previous methods considering the visual and linguistic information in two separate structures, we propose a Visual Language Modeling Network (VisionLAN), which views the visual and linguistic information as a union by directly enduing the vision model with language capability. Specially, we introduce the text recognition of character-wise occluded feature maps in the training stage. Such operation guides the vision model to use not only the visual texture of characters, but also the linguistic information in visual context for recognition when the visual cues are confused (e.g. occlusion, noise, etc.). As the linguistic information is acquired along with visual features without the need of extra language model, VisionLAN significantly improves the speed by 39\% and adaptively considers the linguistic information to enhance the visual features for accurate recognition. Furthermore, an Occlusion Scene Text (OST) dataset is proposed to evaluate the performance on the case of missing character-wise visual cues. The state of-the-art results on several benchmarks prove our effectiveness. Code and dataset are available at \url{https://github.com/wangyuxin87/VisionLAN}.
\end{abstract}

\section{Introduction}

\begin{figure}[t!]
\begin{center}
\includegraphics[width=0.9\linewidth]{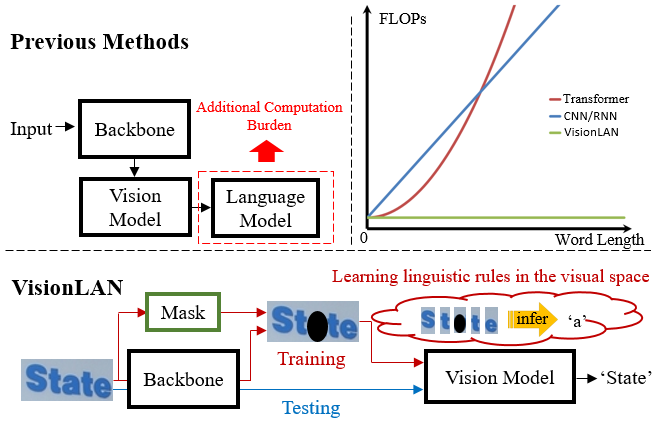}
\end{center}
\vspace{-0.5cm}
\caption{Comparison between previous methods and ours. Top left: the architecture of previous methods. Top right: the extra introduced computation cost for capturing linguistic information when the word length increases. Bottom: the proposed VisionLAN endues the vision model with ability to initiatively capture the linguistic information in visual context during the training stage. In the testing stage, only the vision model is used for prediction.}
\label{fig:1}
\vspace{-0.5cm}
\end{figure}

As a fundamental and pivotal task, scene text recognition (STR) aiming to read the text content from natural images has attracted great interest in computer vision \cite{lee2016recursive,shi2016end,shi2018aster,xie2019aggregation,yue2020robustscanner}. By taking the text image as input and textual prediction as output, some early methods regard the text recognition as a symbol classification task \cite{shi2016end,liao2019scene}. However, it is hard to recognize images with confused visual cues (\emph{e.g.} occlusion, noise, etc.), which are beyond  visual discrimination. As the scene text image contains two-level contents: visual texture and linguistic information, inspired by the Natural Language Processing (NLP) methods \cite{luong2015effective,devlin2018bert}, recent STR works have shifted their research focus to acquiring linguistic information to assist recognition \cite{zhan2019esir,yue2020robustscanner,qiao2020seed,yu2020towards}. Thus, the two-step architecture of vision and language models (top left of Fig. \ref{fig:1}) is popular in recent methods. Specifically, the vision model only focuses on visual texture of characters without considering the linguistic information. Then, the language model predicts the relationship between characters through the linguistic learning structure (RNN \cite{shi2018aster}, CNN \cite{fang2018attention} and Transformer \cite{yu2020towards}).

Though these methods achieve promising results, there are still two problems: 1) \textbf{the extra huge computation cost}. The computation cost of language model increases significantly with the word length getting longer (linear growth for RNN \cite{shi2018aster}/ CNN \cite{fang2018attention} and quadratic growth for Transformer \cite{yu2020towards} in Fig. \ref{fig:1}). Furthermore, many methods adopt a deep bi-directional reasoning architecture \cite{wang2020decoupled,yu2020towards,shi2018aster} to capture more robust linguistic information, which further doubles the computation burden and greatly limits their efficiency in the real application. 2)  \textbf{The difficulty of aggregating two independent information}. It is difficult to comprehensively consider and effectively fuse the visual and linguistic information from two separate structures for accurate recognition \cite{fang2018attention,yue2020robustscanner}. In this paper, we attribute these two problems to the lack of language ability in the vision model, which only focuses on the visual texture of characters without initiatively learning linguistic information \cite{yu2020towards}. As shown in bottom of Fig. \ref{fig:1}, inspired by the human cognitive process that the language capability can be acquired \cite{locke1996infants,harris2006language}, we use vision model as the basic network, and guide it to reason the occluded character during the training stage. Thus, vision model is trained to initiatively learn linguistic information in the visual context. In the test stage, vision model adaptively considers the linguistic information in the visual space for feature enhancement when the visual cues are confused (\emph{e.g.} occlusion, noise, etc.), which effectively supplements the features of occluded characters, and correctly highlights the discriminating visual cues of confused characters (shown in Fig. \ref{fig:vsrm_sg}). To the best of our knowledge, this is \textbf{the first work} to give vision model the ability to perceive language in scene text recognition. We call this new simple architecture as \emph{Visual Language Modeling Network} (VisionLAN).

\begin{figure*}[t!]
\begin{center}
\includegraphics[width=0.85\linewidth]{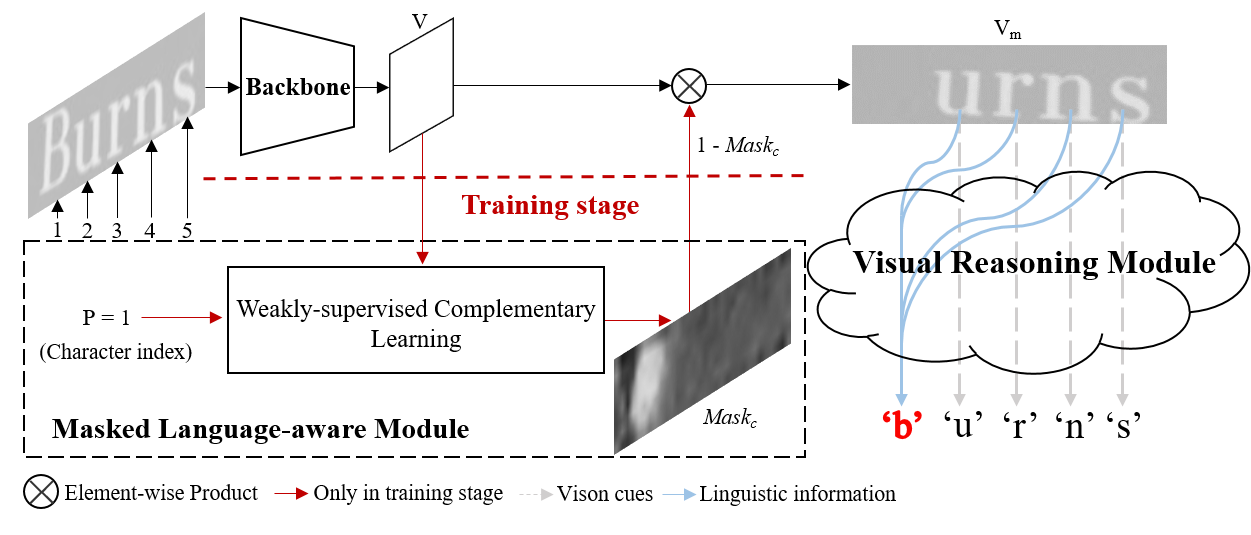}
\end{center}
\vspace{-0.7cm}
\caption{The pipeline of the proposed VisionLAN. VisionLAN mainly contains three parts: backbone network,  \emph{Masked Language-aware Module} (MLM) and  \emph{Visual Reasoning Module} (VRM). MLM is only used in training stage. }
\label{fig:2}
\vspace{-0.5cm}
\end{figure*}

The pipeline of VisionLAN is shown in Fig. \ref{fig:2}. VisionLAN contains three parts: backbone network, \emph{Masked Language-aware Module} (MLM) and \emph{Visual Reasoning Module} (VRM). In the training stage, visual features \emph{V} are firstly extracted from the backbone network. Then MLM takes the visual features \emph{V} and character index \emph{P} as inputs, and automatically generates the character mask map $Mask_c$ at corresponding position through a \emph{Weakly-supervised Complementary Learning}. MLM aims to simulate the case of missing character-wise visual cues by occluding visual messages in \emph{V} with $Mask_c$. In order to consider the linguistic information during the visual texture modeling, we propose a VRM with the ability to capture long-range dependencies in the visual space. VRM takes the occluded feature map $V_m$ as input, and is guided to make the word-level prediction. In the test stage, we remove the MLM and only use VRM  for recognition. As the linguistic information is acquired along with visual features without the need of extra language model, VisionLAN introduces ZERO computation cost for capturing linguistic information (top right in Fig. \ref{fig:1}) and significantly improves the speed by 39\% (Sec. \ref{sec:speed}). Compared with previous methods, the proposed VisionLAN obtains more robust performance on the occluded and low-quality images, and achieves new state-of-the-art results on several benchmarks with a concise pipeline. In addition, an \emph{Occlusion Scene Text} (OST) dataset  is proposed to evaluate the performance on the case of missing character-wise visual cues.

The main contributions of this paper are as follows: 1) A new simple architecture is proposed for scene text recognition. We further visualize the feature maps to illustrate how VisionLAN initiatively uses linguistic information to handle the confused visual cues (\emph{e.g.} occluded, noise, etc.). 2) We propose a \emph{Weakly-supervised Complementary Learning} to generate accurate character-wise mask map in MLM with only word-level annotations. 3) A new \emph{Occlusion Scene Text} (OST) dataset is proposed to evaluate the recognition performance of occluded images. Compared with previous methods, VisionLAN achieves the state-of-the-art performance on seven benchmarks (irregular and regular) and OST with a concise pipeline.

\vspace{-0.3cm}
\section{Related Work}
\vspace{-0.1cm}
\subsection{Scene Text Recognition}
\vspace{-0.1cm}
Scene text recognition (STR) has  been a long-term research topic in computer vision \cite{xie2019aggregation,zhan2019esir,fang2018attention}. With deep learning becoming the most promising machine learning tool \cite{vaswani2017attention,wang2019dsrn,wang2020contournet,ge2021semantic,li2021frequency,lin2021query}, significant progress has been made in the past few years for STR research \cite{qiao2020seed,lyu20192d}. In this section, we divide these methods into two categories  according to whether linguistic rules are used, namely language-free methods and language-aware methods.

Language-free methods \cite{xie2019aggregation,zhang2020adaptive,shi2016end,liao2019scene} view STR as a visual classification task and mainly rely on the visual information for prediction. CRNN \cite{shi2016end} extracts sequential visual features through combined CNN and RNN, then a Connectionist Temporal Classication (CTC) \cite{graves2006connectionist} decoder is used to maximize the probability of all the paths for final prediction. Patel \emph{et al.} \cite{patel2016dynamic} automatically generate the custom lexicon for an image to greatly boost the performance of text reading systems. Zhang \emph{et al.} \cite{zhang2020adaptive} regard text recognition as a visual matching task. They calculate the similarity map between visual features of input image and the pre-defined alphabet to predict the text sequence. Liao \emph{et al.}  \cite{liao2019mask} regard the text recognition as a pixel-wise classification task. Similarly, Textscanner \cite{wan2019textscanner} further proposes an order map to ensure a more accurate transcription from characters to the word. In general, the language-free methods ignore linguistic rules in the recognition process, which usually fail to recognize images with confused visual cues (\emph{e.g.} blur, occlusion, etc.).

Language-aware methods \cite{li2019show,cheng2018aon,zhan2019esir,yang2019symmetry} try to leverage linguistic rules to assist the recognition process. Lee \emph{et al.} \cite{lee2016recursive} use RNNs to automatically learn the sequential dynamics in word strings without manually defining N-grams. Aster \cite{shi2018aster} firstly uses a rectification module before recognition, and then adopts RNNs to model the linguistic information by using the character predicted from the last time step. However, such serial and time-dependent operation in RNN limits the computation efficiency and the performance of semantic reasoning \cite{yu2020towards}. Thus, SRN \cite{yu2020towards} proposes a global semantic reasoning module based on transformer units \cite{vaswani2017attention} for pure language modeling, which takes the prediction of vision model as input and predicts the relationships among characters to refine the recognition results.  Fang \emph{et al.} \cite{fang2018attention} design a completely CNN-based architecture for both vision and language modeling. Though these methods achieve promising results on scene text recognition task, the additionally introduced language model will significantly increase the computation cost. Furthermore, it is also difficult to comprehensively consider and effectively fuse the independent visual and linguistic information in the two-step architecture for accurate recognition \cite{fang2018attention,yue2020robustscanner}. Different from previous methods considering the visual and linguistic information in two separate structures, we directly endue the vision model with language ability and propose a VisionLAN to view the two information as a union. Thus, it is possible to enhance the confused visual cues by capturing linguistic information in the visual context.
\vspace{-0.2cm}
\subsection{Masking and Prediction.}
BERT \cite{devlin2018bert} introduces a cloze task to mask the tokens of input sentence, which is used to learn a robust bi-directional representation based on the context. Following \cite{devlin2018bert}, some works use a similar concept to handle the vision-and-language task \cite{su2019vl,alberti2019fusion,lu2019vilbert}. ViLBERT \cite{lu2019vilbert} uses a two stream model to process visual and textual inputs, and pre-trains their model through the two proxy tasks. Su \emph{et al.} \cite{su2019vl} propose a general structure to fit for most visual-linguistic downstream tasks, which takes both visual and linguistic features as input. As the STR datasets are weakly labeled with word-level annotations, it is difficult to directly implement these masking approaches in STR task. Different from these methods that mask in token or image patch level, in this paper, we propose a \emph{Weakly-supervised Complementary Learning} to automatically mask the input image in the feature level. Thus, VisionLAN learns linguistic information from a new perspective by guiding the model to make word-level prediction on the case of missing character-wise visual cues.
\vspace{-0.2cm}
\section{Proposed Method}
\vspace{-0.1cm}
The VisionLAN is an end-to-end trainable framework with three parts containing: backbone network, \emph{Masked Language-aware Module} (MLM) and \emph{Visual Reasoning Module} (VRM). In this section, we first detail the pipeline of proposed method in Sec. \ref{sec:pipeline}, and then we introduce MLM and VRM in Sec. \ref{sec:pmm} and Sec. \ref{sec:vsrm} respectively.
\vspace{-0.1cm}
\subsection{Pipeline}
\label{sec:pipeline}
\vspace{-0.1cm}
The pipeline of VisionLAN is shown in Fig. \ref{fig:2}. In the training stage, given an input image, the 2D features \emph{V} are firstly extracted from backbone network. Then, MLM takes extracted features $V$ and character index $P$ as inputs, and generates the position-aware character mask map $Mask_c$ through the \emph{Weakly-Supervised Complementary Learning}. $Mask_c$ is used to occlude the character-wise visual messages in \emph{V} to simulate the case of missing character-wise visual semantics. After that, VRM takes occluded feature map $V_m$ as input and makes prediction under the complete word-level supervision. In the testing stage, we remove MLM and only use VRM for prediction.
\vspace{-0.1cm}
\subsection{Masked Language-aware Module}
\label{sec:pmm}
\vspace{-0.1cm}

In order to occlude the character-wise visual cues for the guidance of linguistic learning, we propose a \emph{Masked Language-aware Module} (MLM) to automatically generate the character-wise mask map with only original word-level annotations.

\begin{figure}[t!]
\begin{center}
\includegraphics[width=1.1\linewidth]{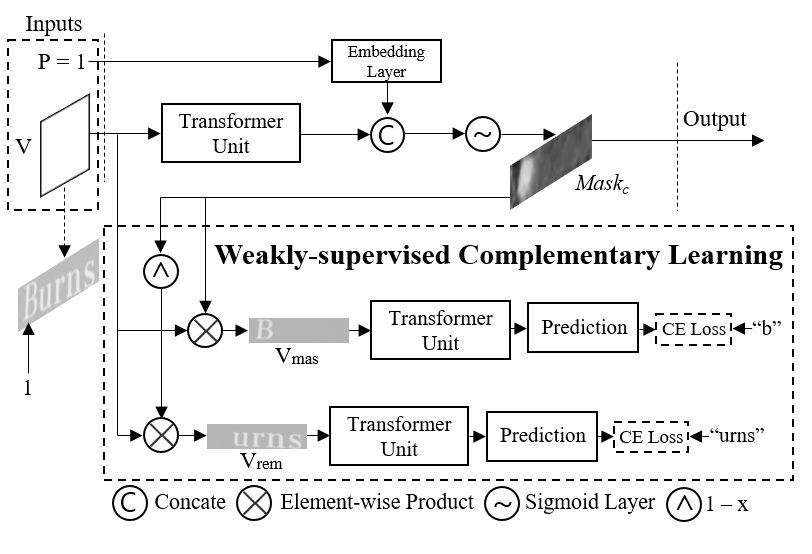}
\end{center}
\vspace{-0.7cm}
\caption{The architecture of MLM. MLM takes the visual features \emph{V} and the character index \emph{P} as inputs to automatically generate character mask map $Mask_c$. CE loss means the cross-entropy loss.}
\label{fig:3}
\vspace{-0.5cm}
\end{figure}

As shown in Fig. \ref{fig:3}, MLM takes visual features \emph{V} and the character index $P$ as inputs. Character index $P$ $\in [1,N_w]$ indicates the index of the occluded character, which is randomly obtained for each input word image with length $N_w$. Then the transformer unit \cite{vaswani2017attention} is used to improve the feature representation ability. Finally, after integrating with character index information, the character mask map $Mask_c$ is obtained through a sigmoid layer, which is used to generate the occluded feature map $V_m$ in Fig. \ref{fig:2}.

To guide the learning process of $Mask_c$, two parallel branches are designed based on the \emph{Weakly-supervised Complementary Learning} (WCL). WCL  aims to guide $Mask_c$ to cover more area of the occluded character, which complementarily makes $1 - Mask_c$ contain more region of other characters. In the first branch, we implement the element-wise product between $V$ and $Mask_c$ to generate the feature map $V_{mas}$ containing visual semantics of the occluded character (\emph{e.g.} character ``b" in the word ``burns" with character index 1 in Fig. \ref{fig:3}). In contrast, the element-wise product between $V$ and $1 - Mask_c$ in the second branch is used to generate the feature map $V_{rem}$ containing visual semantics of other characters (\emph{e.g.} string ``urns" in the word ``burns" in Fig. \ref{fig:3}). By doing these, the complementary learning process guides the $Mask_c$ to only cover the character at corresponding position without overlapping other characters (shown in Fig. \ref{fig:mask}). We share the weights of transformer unit and prediction layer (Eq. \ref{eq:PM_1}) among two parallel branches for the feature representation enhancement and semantic guidance. $V_{in}\in R^{hw\times c}$ is the feature map and $Att\in R^{hw\times N}$ is the attention map, where $c=512$ is the channel number, $N=25$ is the max time step, $h$ and $w$ are the height and width. $O_c$ is positional encoding \cite{vaswani2017attention} of character orders. $W_1$, $W_2$, $W_3$ are trainable weights and $t$ is the time step.

\vspace{-0.3cm}
\begin{equation}
{
   p_{t} = Att_{t}^TV_{in}
}
\label{eq:PM_1}
\end{equation}
\vspace{-0.3cm}
\begin{equation}
{
   Att = Softmax(G(V_{in}))
}
\label{eq:PM_2}
\end{equation}
\vspace{-0.3cm}
\begin{equation}
{
   G(V_{in}) = W_1 tanh(W_2O_c + W_3V_{in})
}
\label{eq:PM_3}
\end{equation}

Compared with BERT \cite{devlin2018bert}, though both approaches mask out the information in a certain time step, the proposed MLM masks the visual features in the 2d spatial space instead of covering token-level information. Furthermore, as STR datasets are weakly labeled, it is difficult to obtain the accurate character-wise pixel-level annotations. Thus, it is impractical to directly implement BERT-based methods \cite{su2019vl,alberti2019fusion,lu2019vilbert} into STR task. Based on these, MLM helps the model to learn linguistic information from a new perspective, which can not be replaced by exiting masking approaches.

The supervisions of WCL are automatically obtained by using the original word-level annotation and randomly generated character index (detailed in Sec. \ref{sec:4}). Thus, MLM automatically generates accurate character mask map without the need of additional annotations, making it possible for the real application.

\subsection{Visual Reasoning Module}
\label{sec:vsrm}

\begin{figure}[t!]
\begin{center}
\includegraphics[width=0.8\linewidth]{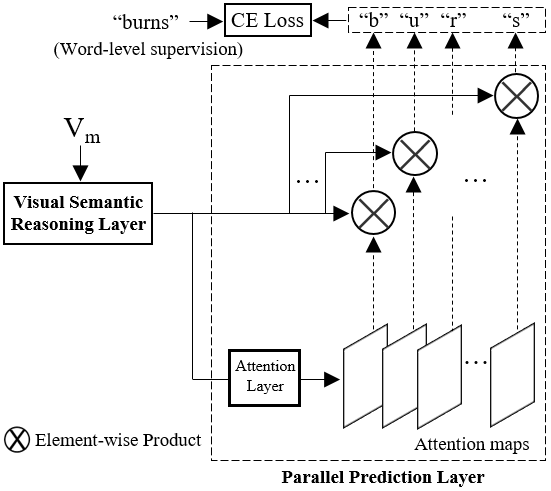}
\end{center}
\vspace{-0.5cm}
\caption{The architecture of VRM. CE loss is cross-entropy loss.}
\label{fig:4}
\vspace{-0.25cm}
\end{figure}

Different from previous methods capturing the visual and linguistic information in the two-step architecture, we propose the \emph{Visual Reasoning Module} (VRM) to model the two information simultaneously in a unified structure. As a pure vision-based structure, VRM aims to reason the word-level prediction from occluded features by using the character-wise information in the visual context.

The details of VRM is shown in Fig. \ref{fig:4}, it contains two parts: Visual Semantic Reasoning (VSR) layer and Parallel Prediction (PP) layer. VSR layer consists of $N$ transformer units \cite{vaswani2017attention}, which are proved to be effective for modeling long-range dependencies in recent computer vision tasks \cite{carion2020end,lyu20192d}. Specially, position encoding is used to perceive the pixel location information. Different from \cite{yu2020towards} using transformer units for pure language modeling, the transformer units in the proposed VRM are used for sequence modeling, which will not be influenced by length of the word. Then, the PP layer is designed to predict the characters in parallel, which has identical formulation as Eq. \ref{eq:PM_1}.

In order to achieve the language modeling process $y_i = f(y_N,...,y_{i+1},y_{i-1},...,y_1)$, the reasoning process of the $i^{th}$ character $y_i$ needs to purely depend on the information of other characters. As MLM accurately occludes the character information in the training stage, VSR layer is guided to predict the dependencies between visual features of characters to infer the semantics of occluded character. Thus, with the word-level supervision, VSR layer learns to initiatively model the linguistic information in visual context to assist recognition. In the testing stage, VSR layer is able to adaptively consider the linguistic information for visual feature enhancement when the current visual semantics are confused (\emph{e.g.} occlusion, noise, etc.).

\begin{figure}[t!]
\begin{center}
\includegraphics[width=0.85\linewidth]{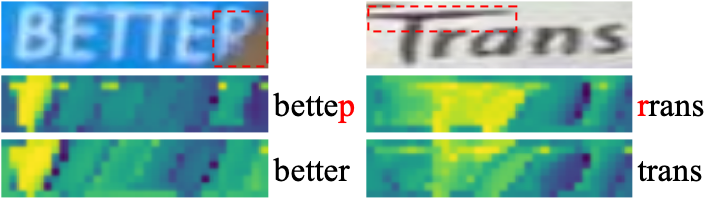}
\end{center}
\vspace{-0.5cm}
\caption{The visualization of features generated from VSR layer and the corresponding prediction result. Top: input image. Middle: model implemented without MLM. Bottom: our VisionLAN.}
\label{fig:vsrm_sg}
\vspace{-0.5cm}
\end{figure}

We visualize the feature maps generated from VSR layer in testing to better understand how the learned linguistic information improves the recognition performance. As shown in Fig. \ref{fig:vsrm_sg}, VSR layer effectively supplements the semantics of occluded character ``r" in the word ``better", and correctly highlights the discriminating visual cues of character ``t" in the word ``trans" with the help of linguistic information in visual context. Without the initiative linguistic learning guided by MLM, VRM wrongly predicts the input images as ``bettep" and ``rrans".

\subsection{Training Objective}

The final objective function of the proposed method is formulated in Eq. \ref{eq:2}. $L_{rec}$ is loss in VRM, and $L_{mas}$ \& $L_{rem}$ are losses for predicting masked character and other characters in MLM respectively. $\lambda_{1}$ and $\lambda_{2}$ are used to balance the losses. Specially, we set $\lambda_{1} = \lambda_{2} = 0.5$, and use cross-entropy loss formulated in Eq. \ref{eq:1} for $L_{rec}$, $L_{mas}$ and $L_{rem}$. $p_t$ and $g_t$ represent the prediction and ground truth. We set N to 25 in our experiments.
\begin{equation}
{
    L = L_{rec} + \lambda_{1} L_{mas} + \lambda_{2} L_{rem}
}
\label{eq:2}
\end{equation}
\begin{equation}
{
   L_{*} = -\frac{1}{N}\sum_{t=1}^N\log(p_t|g_t)
}
\label{eq:1}
\end{equation}

\section{Experiment}
\label{sec:4}

\subsection{Datasets}

We conduct experiments following the setup of \cite{yu2020towards} in the purpose of fair comparison. The training datasets are SynthText (ST) \cite{gupta2016synthetic} and SynthText90K (90K) \cite{jaderberg2014synthetic}. The performance is evaluated on 6 benchmarks containing IIIT 5K-Words (IIIT5K) \cite{mishra2012scene}, ICDAR2013 (IC13) \cite{karatzas2013icdar}, ICDAR2015 (IC15) \cite{karatzas2015icdar}, Street View Text (SVT) \cite{wang2011end}, Street View Text-Perspective (SVTP) \cite{quy2013recognizing} and CUTE80 (CT) \cite{risnumawan2014robust}. Details of above 6 datasets can be found in previous works \cite{yu2020towards,qiao2020seed}.

\begin{figure}[t!]
\begin{center}
\includegraphics[width=0.8\linewidth]{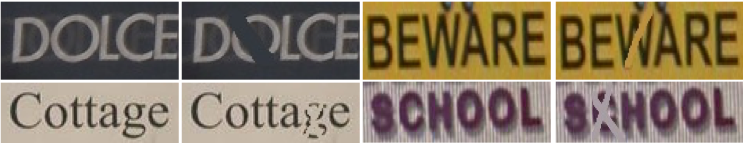}
\end{center}
\vspace{-0.3cm}
\caption{The examples of text images in OST. Top: image occluded in weak degree. Bottom: image occluded in heavy degree. Left: original image. Right: occluded image.}
\label{fig:ost}
\vspace{-0.2cm}
\end{figure}

In addition, we provide a new Occlusion Scene Text (OST) dataset to reflect the ability for recognizing cases with missing visual cues. This dataset is collected from 6 benchmarks (IC13, IC15, IIIT5K, SVT, SVTP and CT) containing 4832 images. Images in this dataset are manually occluded in weak or heavy degree (shown in Fig. \ref{fig:ost}). Weak and heavy degrees mean that we occlude the character using one or two lines. For each image, we randomly choose one degree to only cover one character. More examples of OST are shown in the supplementary materials.

\vspace{-0.2cm}
\subsection{Implementation Details}
\vspace{-0.1cm}
We use the ResNet45 \cite{shi2018aster,wang2020decoupled,qiao2020seed} as our backbone. Particularly, we set the stride to 2 in stage 2,3,4 and initialize the weights by default. Following the most recent works \cite{yu2020towards,qiao2020seed}, we set the image size to $256 \times 64$ (there is no obvious difference with the size of $128 \times 32$ in our experiments). Data augmentation including random rotation, color jittering and perspective distortion. We conduct the experiments on 4 NVIDIA V100 GPUs with batch size 384. The network is trained end-to-end using Adam optimizer with learning rate 1e-4. The recognition covers 37 characters including a-z, 0-9, and an end-of-sequence symbol.

Following \cite{yu2020towards}, we divide the training process into 2 steps: language-free (LF) step and language-aware (LA) step. It is worth mentioning that we control the total number of training sessions to be consistent with existing methods for fair comparison. 1) In LF step, we split the connection between MLM and VRM ($V = V_m$ in Fig. \ref{fig:2}) to guarantee a more stable learning process of both modules. VRM in this step will not acquire the language capability and only uses visual texture for prediction. 2) In LA step, $Mask_c$ generated from MLM is used to occlude the feature map $V$ to guide the learning of linguistic rules in VRM. Specifically, we control the ratio of occluded number in a batch, which aims to balance the cases with rich or weak visual information during the training stage.

As all the training images have word-level annotations, we randomly generate the character index based on the length of word, and use this index and the original word-level annotation to generate the labels for MLM (\emph{e.g.} when index is 4 and word is ``house", the labels are ``s" and ``houe" respectively). The label generating process is automatic without manual intervention, making it easy to finetune our model on other datasets.
\vspace{-0.2cm}
\subsection{Ablation Study}
We illustrate the effectiveness of proposed modules in this section. To be specific, baseline contains VRM with two transformer units in Tab. \ref{tab:1}\& \ref{tab:2}\& \ref{tab:dropout}.
\vspace{-0.1cm}
\begin{table}[t!]
\caption{Ablation study about the occluded number ratio of one batch in the MLM during the training stage.}
\vspace{-0.5cm}
\begin{center}
\begin{tabular}{|l|c|c|c|c|c|c|}
\hline
Ratio&IIIT5K&IC13&SVT&IC15&SVTP&CT \\
\hline
Baseline & 94.5& 94.2& 89.3 & 79.8 & 81.1 & 85.8 \\
1:2 & 95.0& 94.8&90.4& 80.8& 83.0& 88.0 \\
1:1 & \textbf{95.4}& \textbf{95.0}&  \textbf{91.0}& \textbf{81.8}& \textbf{83.7}& \textbf{88.2}\\
2:1 & 95.0 & 94.7& 90.0& 81.1& 82.7& 88.1\\
\hline
\end{tabular}
\end{center}
\label{tab:1}
\vspace{-0.5cm}
\end{table}
\begin{table}[t!]
\caption{Ablation study of the WCL. ``Mas only" and ``Rem only" means that we only implement the $1^{st}$ or the $2^{nd}$ branch in MLM.}
\vspace{-0.6cm}
\begin{center}
\begin{tabular}{|l|c|c|c|c|c|c|}
\hline
Methods&IIIT5K&IC13&SVT&IC15&SVTP&CT \\
\hline
Mas only & 94.8 & 94.7& 89.8& 81.7& 82.3& 87.2\\
Rem only & 95.2 & 94.8& 89.9& 81.1& 82.2& 88.0 \\
WCL& \textbf{95.4}& \textbf{95.0}&  \textbf{91.0}& \textbf{81.8}& \textbf{83.7}& \textbf{88.2}\\
\hline
\end{tabular}
\end{center}
\label{tab:2}
\vspace{-0.5cm}
\end{table}
\begin{table}[t!]
\caption{The comparisons between MLM and other masking methods. Average accuracy is calculated from 6 benchmarks. We set cutout patch to $h\times w/10$ and dropout value to 0.1. The results are compared under the same training sessions.}
\vspace{-0.3cm}
\begin{center}
\begin{tabular}{|l|c|}
\hline
Methods& Average accuracy(\%)\\
\hline
Baseline & 88.8\\
Dropout \cite{srivastava2014dropout}& 89.0\\
Cutout \cite{devries2017improved}& 89.0\\
MLM & \textbf{90.2}\\
\hline
\end{tabular}
\end{center}
\label{tab:dropout}
\vspace{-0.5cm}
\end{table}
\begin{table}[t!]
\caption{Ablation study about the ability of linguistic information capturing in VRM. ``2L" means two transformer units are used.}
\vspace{-0.6cm}
\begin{center}
\begin{tabular}{|l|c|c|c|c|c|c|}
\hline
Methods&IIIT5K&IC13&SVT&IC15&SVTP&CT \\
\hline
VRM-2L & 95.4& 95.0& 91.0& 81.8 & 83.7 & 88.2 \\
VRM-3L & \textbf{95.8}& \textbf{95.7} &  \textbf{91.7} & \textbf{83.7} &  \textbf{86.0} & \textbf{88.5} \\
\hline
\end{tabular}
\end{center}
\label{tab:3}
\vspace{-0.7cm}
\end{table}

\vspace{-0.15cm}
~\\
\noindent\textbf{The effectiveness of MLM.} The proposed MLM aims to guide the linguistic learning process in the VRM. We conduct several experiments to evaluate its effectiveness in Tab. \ref{tab:1}. The baseline model is implemented without MLM. We change the ratio of occluded number in a batch to study its influence to the recognition performance  (\emph{e.g.}  when the batch size is 128, $ratio$ = 1:3 means that we use $Mask_c$ to occlude V for only 32 samples in 1 batch, and feature maps of the rest 96 samples remain unchanged). As shown in Tab. \ref{tab:1}, the proposed MLM significantly improves the performance of baseline model when the ratio ranges from 1:2 to 2:1. For the irregular datasets (IC15, SVTP, CT) containing amounts of images with confused visual cues (\emph{blur, occlusion, noise, etc.}), the proposed MLM improves the baseline model at least 2\% in accuracy with $ratio = $1:1, which further demonstrates that the initiative linguistic learning process effectively helps the vision model to handle confused visual cues. For regular datasets, the improvement is also considerable (0.9\%, 0.8\%, and 1.7\% on IIIT5K, IC13 and SVT datasets respectively). When the ratio raises up to 2:1, the performance drops slightly. We infer that the large value of ratio will break the balance between cases with rich and weak visual cues during the training process. Therefore, we set the value of ratio to 1:1 in the rest experiments.

\begin{table*}[t!]
\caption{Results on IIIT5K, IC13, SVT, IC15, SVTP and CUTE datasets. Following \cite{qiao2020seed,yu2020towards}, all the results are under NONE lexicon. Lan-free and Lan-aware are shorts for language-free and language-aware methods. ``Annos" is short for annotations. ``char" and ``word" mean character-level and word-level annotations are used in the training stage. Baseline contains VRM with three transformer units. }
\vspace{-0.3cm}
\begin{center}
\begin{tabular}{|c|l|c|c|c|c|c|c|c|c|}
\hline
 & Methods&Training Data&Annos&IIIT5K&IC13&SVT&IC15&SVTP&CT\\
\hline
\multirow{3}*{Lan-free}
&CTC \cite{shi2016end}&90K&word & 81.2& 89.6& 82.7& -& -& -\\
&ACE \cite{xie2019aggregation} &90K&word & 82.3 & 89.7 & 82.6 & 68.9 & 70.1 & 82.6 \\
&FCN \cite{liao2019scene}&ST&word, char & 91.9& 91.5& 86.4& -& -&- \\
\hline
\multirow{11}*{Lan-aware}
&FAN \cite{cheng2017focusing} &90K+ST&word & 87.4& 93.3& 85.9& 70.6& - & -\\
&AON \cite{cheng2018aon} &90K+ST&word & 87.0& -& 82.8& 68.2& 73.0& 76.8 \\
&ASTER \cite{shi2018aster}&90K+ST&word & 93.4& 91.8& 89.5& 76.1& 78.5& 79.5  \\
&ESIR \cite{zhan2019esir} &90K+ST&word & 93.3& 91.3& 90.2& 76.9& 79.6& 83.3  \\
 &ScRN \cite{yang2019symmetry} &90K+ST&word, char & 94.4 & 93.9 & 88.9 & 78.7 & 80.8 & 87.5  \\
&SAR \cite{li2019show} &90K+ST&word & 91.5 & 91.0 & 84.5 & 69.2 & 76.4& 83.3 \\
&TextScanner \cite{wan2019textscanner}& 90K+ST & word, char & 83.9 & 92.9 & 90.1 & 79.4 & 84.3 & 83.3  \\
& DAN \cite{wang2020decoupled} &90K+ST&word & 94.3& 93.9& 89.2& 74.5& 80.0& 84.4\\
&Wang \emph{et al.} \cite{wang2020exploring} & 90K+ST & word & 94.4 & 93.7 & 89.8 & 75.1 & 80.2 & 86.8 \\
&SRN \cite{yu2020towards} &90K+ST&word & 94.8 & 95.5 & 91.5 & 82.7 & 85.1 & 87.8 \\
&SEED \cite{qiao2020seed}&90K+ST&word & 93.8 & 92.8 & 89.6 & 80.0 & 81.4 & 83.6 \\
\hline
\multirow{2}*{Ours}
&Baseline&90K+ST&word & 94.6& 94.3& 89.3 & 81.2 & 81.6 & 86.8  \\
&\textbf{VisionLAN} &90K+ST&word & \textbf{95.8}& \textbf{95.7} &  \textbf{91.7} & \textbf{83.7} &  \textbf{86.0} & \textbf{88.5} \\
\hline
\end{tabular}
\end{center}
\label{tab:4}
\vspace{-0.85cm}
\end{table*}

\vspace{-0.2cm}
~\\
\noindent\textbf{The effectiveness of WCL.} To demonstrate the effectiveness of proposed \emph{Weakly-supervised Complementary Learning} in MLM, we conduct several experiments implemented with only the first branch (occluded character) or the second branch (remaining string). As shown in Tab. \ref{tab:2}, MLM implemented with the complementary learning process obtains better results than the methods only guiding the semantics of occluded character or remaining string during the training stage.

\vspace{-0.2cm}
~\\
\noindent\textbf{Compared with other masking methods.} We compare MLM with \cite{devries2017improved,srivastava2014dropout} to evaluate our effectiveness in language modeling. All the modules only work on $V$ for fair comparison. As shown in Tab. \ref{tab:dropout}, the proposed MLM significantly improves the recognition results (1.4\% vs 0.2\%). As detailed in Sec. \ref{sec:vsrm}, the reasoning process of the $i^{th}$ character needs to purely depend on the information of other characters without containing current character-wise information. Thus, randomly masking pixel-wise feature \cite{devries2017improved,srivastava2014dropout} does not have the ability of linguistic learning.  Benefiting from the well-designed architecture and ingenious weakly supervised learning, MLM accurately localizes character-wise visual cues, which has the ability to guide the linguistic learning process in VRM.

\vspace{-0.2cm}
~\\
\noindent\textbf{The effectiveness of VRM.} To study the relationship between the recognition performance and the ability of capturing linguistic information, we compare the results of models implemented with different number of transformer units in VSR layer. As shown in Tab. \ref{tab:3}, VRM implemented with three transformer units further improves the performance, which has the stronger language capability.
\subsection{Comparisons with State-of-the-Arts}
\label{sec:speed}
We compare our method with previous state-of-the-art methods on 6 benchmarks in Tab. \ref{tab:4}. We simply divide the methods into language-free and language-aware methods according to whether linguistic information are used. The language-aware methods perform better than language-free methods in general. Benefiting from adaptively considering the linguistic information for feature enhancement, the proposed VisionLAN achieves state-of-the-art performance across the 6 public datasets compared with both language-free and language-aware methods. Specifically, for regular datasets, the proposed VisionLAN obtains 1\%, 0.2\% and 0.2\% improvement on IIIT5K, IC13 and SVT datasets respectively. For irregular datasets, the increases are  1\%, 0.9\% and 0.7\% on IC15, SVTP and CT respectively.

\begin{table}[t!]
\caption{The comparisons of speed and EIPs between existing language models and ours. The test dataset is IC15. }
\vspace{-0.2cm}
\begin{center}
\begin{tabular}{|l|c|c|c|c|}
\hline
Methods& Speed& EIPs\\
\hline
Baseline & \textbf{11.5ms}& - \\
Baseline + \cite{shi2018aster}& 43.2ms& 3.0M\\
Baseline + \cite{yu2020towards} & 19ms& 12.6M\\
VisionLAN & \textbf{11.5ms}&\textbf{0M} \\
\hline
\end{tabular}
\end{center}
\label{tab:eips}
\vspace{-0.7cm}
\end{table}
\begin{table}[t!]
\caption{Results on OST. Average is the short for average accuracy. Weak and Heavy mean the accuracy on weak and heavy degrees.}
\vspace{-0.3cm}
\begin{center}
\begin{tabular}{|l|c|c|c|}
\hline
Methods& Average& Weak& Heavy\\
\hline
Baseline & 53.0& 63.2& 42.7\\
Baseline + \cite{shi2018aster} & 53.9 & 63.9& 43.9\\
Baseline + \cite{yu2020towards} & 58.2& 68.4& 48.0\\
VisionLAN & \textbf{60.3}& \textbf{70.3}& \textbf{50.3}\\
\hline
\end{tabular}
\end{center}
\label{tab:ost}
\vspace{-0.7cm}
\end{table}

As VisionLAN adaptively considers the visual and linguistic information in the 2d visual space, our method is less sensitive to the distorted images. Thus, the proposed method can obtain better results than ASTER \cite{shi2018aster} and ESIR \cite{zhan2019esir} on irregular datasets, which adopt the rectification process before recognition. As shown in Tab. \ref{tab:4}, the increases are 7.6\%, 7.5\% and 9\% for \cite{shi2018aster}, and 6.8\%, 6.4\% and 5.2\% for \cite{zhan2019esir}  on IC15, SVTP and CT datasets respectively.

We further compare the differences between the existing methods and ours in recognition speed and the extra introduced parameters (EIPs) for capturing linguistic information in Tab. \ref{tab:eips}. In terms of approaching speed and parameters, we implement one transformer unit in GSRM of \cite{yu2020towards} (the same goes for Sec. \ref{sec:ost_eval}). As the linguistic information is acquired along with the visual features without the need of extra language model, the proposed VisionLAN significantly improves the speed by at least 39\%  (11.5ms vs 19ms and 43.2ms) without introducing extra parameters (0M vs 12.6M and 3M). Furthermore, as VisionLAN directly considers the linguistic information in the visual space, its efficiency of capturing linguistic information will not be affected by the word length.

\vspace{-0.1cm}
\subsection{The Language Capability on OST Dataset}
\label{sec:ost_eval}
\vspace{-0.1cm}

To evaluate the language capability of our VisionLAN in detail, we compare our method with recent most popular language models (RNN \cite{shi2018aster}  and Transformer \cite{yu2020towards}) on OST dataset to evaluate their performance on the case of missing character-wise visual cues. Specifically, we connect these language models to VRM following the implementation details in their papers. As shown in Tab. \ref{tab:ost}, though the linguistic information captured by \cite{shi2018aster} and \cite{yu2020towards} can assist the prediction of vision model, the proposed VisionLAN significantly outperforms these methods by viewing the visual and linguistic information as a union. Through adaptively aggregating the two information in a unified structure instead of considering them independently, VisionLAN improves the baseline model by 7.3\% in average.

\vspace{-0.1cm}
\subsection{The Generalization Ability on Long Chinese Dataset}
\vspace{-0.1cm}
We evaluate VisionLAN on non-Latin Long Text (TRW15 \cite{zhou2015icdar}) to prove its generalization ability. This dataset contains 2997 cropped images, and we set the max length $N$ to 50. We train the proposed VisionLAN following the setup of \cite{yu2020towards}. As shown in Tab. \ref{tab:TRW}, compared with language-free (CTC) and language-aware (2D-Attention) methods, VisionLAN outperforms these approaches by at least 14.9\%. Benefiting from viewing the visual and linguistic information as a union, the proposed VisionLAN achieves a new state-of-the-art result and significantly outperforms SRN \cite{yu2020towards} by 3.2\%. More experiments on other datasets (\emph{e.g.} MLT \cite{nayef2019icdar2019}, etc.) are available in the supplementaries.
\begin{table}[t!]
\caption{Results on TRW15 \cite{zhou2015icdar}.}
\begin{center}
\begin{tabular}{|l|c|}
\hline
Methods& Accuracy(\%)\\
\hline
SCCM \cite{yin2017scene} & 81.2\\
2D-Attention \cite{yu2020towards} & 72.2\\
CTC \cite{yu2020towards} & 73.8\\
SRN \cite{yu2020towards} & 85.5\\
VisionLAN & \textbf{88.7}\\
\hline
\end{tabular}
\end{center}
\label{tab:TRW}
\vspace{-0.8cm}
\end{table}

\subsection{The Qualitative Analysis}

\begin{figure}[t!]
\begin{center}
\includegraphics[width=0.7\linewidth]{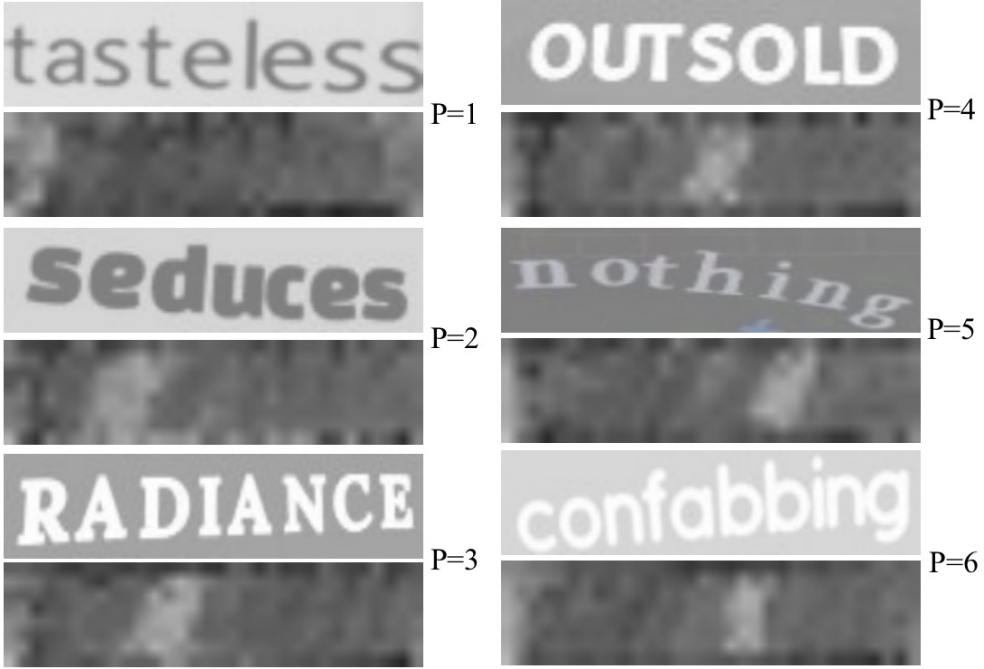}
\end{center}
\vspace{-0.4cm}
\caption{The examples of generated $Mask_c$. The top image is input image, and the bottom image is $Mask_c$ with corresponding character index $P$.}
\label{fig:mask}
\vspace{-0.2cm}
\end{figure}

\noindent\textbf{MLM in character-wise localization.} To qualitatively analyze the effectiveness of MLM, we visualize some examples of generated $Mask_c$ in Fig. \ref{fig:mask}. The generated $Mask_c$ effectively localizes character-wise visual cues at corresponding position with the guidance of character index $P$. Furthermore, MLM is able to handle the distorted images (\emph{e.g.} the curved word image ``nothing") and the localization of repeated characters (\emph{e.g.} the character ``b" with $P=6$ in word ``confabbing"). The quantitative evaluation of character-wise localization performance and more visualizations of $Mask_c$ are available in the supplementaries.

\begin{figure}[t!]
\begin{center}
\includegraphics[width=0.85\linewidth]{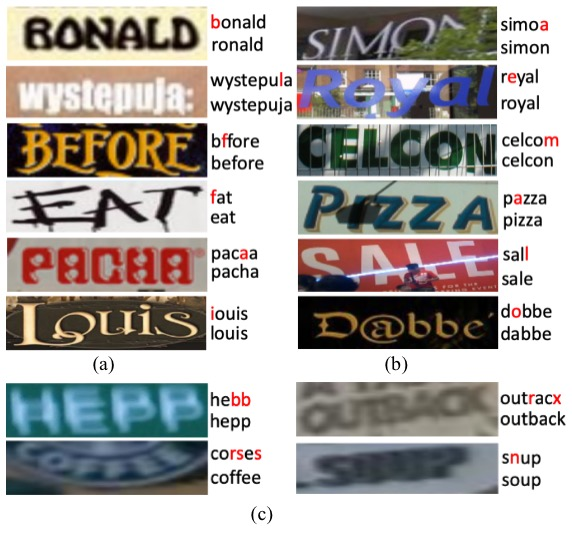}
\end{center}
\vspace{-0.5cm}
\caption{Recognition results with/without the use of linguistic information. The top string is the prediction of VisionLAN without MLM. The bottom string is the prediction of VisionLAN. (a): characters with confused visual cues. (b): characters disturbed by background. (c): blurred characters.}
\label{fig:vlm}
\vspace{-0.6cm}
\end{figure}

\vspace{-0.2cm}
~\\
\noindent\textbf{The effectiveness of VisionLAN.} We collect some recognition results to illustrate how the learned linguistic information helps vision model to improve the performance. As shown in  Fig. \ref{fig:vlm} (a), VisionLAN can handle the cases with confusing characters. For example, as the character ``e" has the similar visual cues to character ``f" in the image with word ``before", the VisionLAN without MLM wrongly gives the prediction ``f", while VisionLAN correctly infers the character ``e" with the help of linguistic information. For the samples in Fig. \ref{fig:vlm} (b), VisionLAN can also use linguistic rules to eliminate the background interference (including occlusion, illumination, background textures, etc.). Furthermore, the accurate recognition of the blurred characters in Fig. \ref{fig:vlm} (c) also demonstrates the effectiveness of our method.
\vspace{-0.4cm}
\section{Conclusion}
\vspace{-0.2cm}
As the first work to endue the vision model with language capability, this paper proposes a concise and effective architecture for scene text recognition. VisionLAN successfully achieves the transformation from two-step to one-step recognition (from Two to One), which adaptively considers both visual and linguistic information in a unified structure without the need of  extra language model. Compared with previous language model, VisionLAN shows a stronger language capability while maintaining high efficiency. In addition, a new \emph{Occlusion Scene Text} dataset is proposed to evaluate the performance on the cases of missing character-wise visual cues. Extensive experiments on seven benchmarks and the proposed OST dataset demonstrate the effectiveness and efficiency of our method. We regard the proposed VisionLAN as a basic step toward more robust and accurate scene text recognition, and we will further explore its potential in the future.


\section*{Acknowledgments}
This work is supported by the National Nature Science Foundation of China (62121002, 62022076, U1936210), the Fundamental Research Funds for the Central Universities under Grant WK3480000011, the China Postdoctoral Science Foundation (2021M693092) and JD AI research. We also acknowledge the support of GPU cluster built by MCC Lab of Information Science and Technology Institution, USTC.

{\small
\bibliographystyle{ieee_fullname}
\bibliography{egbib}

\begin{thebibliography}{10}\itemsep=-1pt

\bibitem{alberti2019fusion}
Chris Alberti, Jeffrey Ling, Michael Collins, and David Reitter.
\newblock Fusion of detected objects in text for visual question answering.
\newblock In {\em Proceedings of the 2019 Conference on Empirical Methods in
  Natural Language Processing and the 9th International Joint Conference on
  Natural Language Processing (EMNLP-IJCNLP)}, pages 2131--2140, 2019.

\bibitem{carion2020end}
Nicolas Carion, Francisco Massa, Gabriel Synnaeve, Nicolas Usunier, Alexander
  Kirillov, and Sergey Zagoruyko.
\newblock End-to-end object detection with transformers.
\newblock {\em arXiv preprint arXiv:2005.12872}, 2020.

\bibitem{cheng2017focusing}
Zhanzhan Cheng, Fan Bai, Yunlu Xu, Gang Zheng, Shiliang Pu, and Shuigeng Zhou.
\newblock Focusing attention: Towards accurate text recognition in natural
  images.
\newblock In {\em Proceedings of the IEEE international conference on computer
  vision}, pages 5076--5084, 2017.

\bibitem{cheng2018aon}
Zhanzhan Cheng, Yangliu Xu, Fan Bai, Yi Niu, Shiliang Pu, and Shuigeng Zhou.
\newblock Aon: Towards arbitrarily-oriented text recognition.
\newblock In {\em Proceedings of the IEEE Conference on Computer Vision and
  Pattern Recognition}, pages 5571--5579, 2018.

\bibitem{devlin2018bert}
Jacob Devlin, Ming-Wei Chang, Kenton Lee, and Kristina Toutanova.
\newblock Bert: Pre-training of deep bidirectional transformers for language
  understanding.
\newblock {\em arXiv preprint arXiv:1810.04805}, 2018.

\bibitem{devries2017improved}
Terrance DeVries and Graham~W Taylor.
\newblock Improved regularization of convolutional neural networks with cutout.
\newblock {\em arXiv preprint arXiv:1708.04552}, 2017.

\bibitem{fang2018attention}
Shancheng Fang, Hongtao Xie, Zheng-Jun Zha, Nannan Sun, Jianlong Tan, and
  Yongdong Zhang.
\newblock Attention and language ensemble for scene text recognition with
  convolutional sequence modeling.
\newblock In {\em Proceedings of the 26th ACM international conference on
  Multimedia}, pages 248--256, 2018.

\bibitem{ge2021semantic}
Jiannan Ge, Hongtao Xie, Shaobo Min, and Yongdong Zhang.
\newblock Semantic-guided reinforced region embedding for generalized zero-shot
  learning.
\newblock In {\em Proceedings of the AAAI Conference on Artificial
  Intelligence}, volume~35, pages 1406--1414, 2021.

\bibitem{graves2006connectionist}
Alex Graves, Santiago Fern{\'a}ndez, Faustino Gomez, and J{\"u}rgen
  Schmidhuber.
\newblock Connectionist temporal classification: labelling unsegmented sequence
  data with recurrent neural networks.
\newblock In {\em Proceedings of the 23rd international conference on Machine
  learning}, pages 369--376, 2006.

\bibitem{gupta2016synthetic}
Ankush Gupta, Andrea Vedaldi, and Andrew Zisserman.
\newblock Synthetic data for text localisation in natural images.
\newblock In {\em Proceedings of the IEEE conference on computer vision and
  pattern recognition}, pages 2315--2324, 2016.

\bibitem{harris2006language}
Catherine~L Harris.
\newblock Language and cognition.
\newblock {\em Encyclopedia of cognitive science}, pages 1--6, 2006.

\bibitem{jaderberg2014synthetic}
Max Jaderberg, Karen Simonyan, Andrea Vedaldi, and Andrew Zisserman.
\newblock Synthetic data and artificial neural networks for natural scene text
  recognition.
\newblock {\em NIPS}, 2014.

\bibitem{karatzas2015icdar}
Dimosthenis Karatzas, Lluis Gomez-Bigorda, Anguelos Nicolaou, Suman Ghosh,
  Andrew Bagdanov, Masakazu Iwamura, Jiri Matas, Lukas Neumann,
  Vijay~Ramaseshan Chandrasekhar, Shijian Lu, et~al.
\newblock Icdar 2015 competition on robust reading.
\newblock In {\em 2015 13th International Conference on Document Analysis and
  Recognition (ICDAR)}, pages 1156--1160. IEEE, 2015.

\bibitem{karatzas2013icdar}
Dimosthenis Karatzas, Faisal Shafait, Seiichi Uchida, Masakazu Iwamura,
  Lluis~Gomez i Bigorda, Sergi~Robles Mestre, Joan Mas, David~Fernandez Mota,
  Jon~Almazan Almazan, and Lluis~Pere De~Las~Heras.
\newblock Icdar 2013 robust reading competition.
\newblock In {\em 2013 12th International Conference on Document Analysis and
  Recognition}, pages 1484--1493. IEEE, 2013.

\bibitem{lee2016recursive}
Chen-Yu Lee and Simon Osindero.
\newblock Recursive recurrent nets with attention modeling for ocr in the wild.
\newblock In {\em Proceedings of the IEEE Conference on Computer Vision and
  Pattern Recognition}, pages 2231--2239, 2016.

\bibitem{li2019show}
Hui Li, Peng Wang, Chunhua Shen, and Guyu Zhang.
\newblock Show, attend and read: A simple and strong baseline for irregular
  text recognition.
\newblock In {\em Proceedings of the AAAI Conference on Artificial
  Intelligence}, volume~33, pages 8610--8617, 2019.

\bibitem{li2021frequency}
Jiaming Li, Hongtao Xie, Jiahong Li, Zhongyuan Wang, and Yongdong Zhang.
\newblock Frequency-aware discriminative feature learning supervised by
  single-center loss for face forgery detection.
\newblock In {\em Proceedings of the IEEE/CVF Conference on Computer Vision and
  Pattern Recognition}, pages 6458--6467, 2021.

\bibitem{liao2019mask}
Minghui Liao, Pengyuan Lyu, Minghang He, Cong Yao, Wenhao Wu, and Xiang Bai.
\newblock Mask textspotter: An end-to-end trainable neural network for spotting
  text with arbitrary shapes.
\newblock {\em IEEE transactions on pattern analysis and machine intelligence},
  2019.

\bibitem{liao2019scene}
Minghui Liao, Jian Zhang, Zhaoyi Wan, Fengming Xie, Jiajun Liang, Pengyuan Lyu,
  Cong Yao, and Xiang Bai.
\newblock Scene text recognition from two-dimensional perspective.
\newblock In {\em Proceedings of the AAAI Conference on Artificial
  Intelligence}, volume~33, pages 8714--8721, 2019.

\bibitem{lin2021query}
Fanchao Lin, Hongtao Xie, Yan Li, and Yongdong Zhang.
\newblock Query-memory re-aggregation for weakly-supervised video object
  segmentation.
\newblock In {\em Proceedings of the AAAI Conference on Artificial
  Intelligence}, volume~35, pages 2038--2046, 2021.

\bibitem{locke1996infants}
John~L Locke.
\newblock Why do infants begin to talk? language as an unintended consequence.
\newblock {\em Journal of child language}, 23(2):251--268, 1996.

\bibitem{lu2019vilbert}
Jiasen Lu, Dhruv Batra, Devi Parikh, and Stefan Lee.
\newblock Vilbert: Pretraining task-agnostic visiolinguistic representations
  for vision-and-language tasks.
\newblock In {\em Advances in Neural Information Processing Systems}, pages
  13--23, 2019.

\bibitem{luong2015effective}
Minh-Thang Luong, Hieu Pham, and Christopher~D Manning.
\newblock Effective approaches to attention-based neural machine translation.
\newblock In {\em Proceedings of the 2015 Conference on Empirical Methods in
  Natural Language Processing}, pages 1412--1421, 2015.

\bibitem{lyu20192d}
Pengyuan Lyu, Zhicheng Yang, Xinhang Leng, Xiaojun Wu, Ruiyu Li, and Xiaoyong
  Shen.
\newblock 2d attentional irregular scene text recognizer.
\newblock {\em arXiv preprint arXiv:1906.05708}, 2019.

\bibitem{mishra2012scene}
Anand Mishra, Karteek Alahari, and CV Jawahar.
\newblock Scene text recognition using higher order language priors.
\newblock In {\em BMVC}, 2012.

\bibitem{nayef2019icdar2019}
Nibal Nayef, Yash Patel, Michal Busta, Pinaki~Nath Chowdhury, Dimosthenis
  Karatzas, Wafa Khlif, Jiri Matas, Umapada Pal, Jean-Christophe Burie,
  Cheng-lin Liu, et~al.
\newblock Icdar2019 robust reading challenge on multi-lingual scene text
  detection and recognition—rrc-mlt-2019.
\newblock In {\em 2019 International Conference on Document Analysis and
  Recognition (ICDAR)}, pages 1582--1587. IEEE, 2019.

\bibitem{patel2016dynamic}
Yash Patel, Lluis Gomez, Mar{\c{c}}al Rusinol, and Dimosthenis Karatzas.
\newblock Dynamic lexicon generation for natural scene images.
\newblock In {\em European Conference on Computer Vision}, pages 395--410.
  Springer, 2016.

\bibitem{qiao2020seed}
Zhi Qiao, Yu Zhou, Dongbao Yang, Yucan Zhou, and Weiping Wang.
\newblock Seed: Semantics enhanced encoder-decoder framework for scene text
  recognition.
\newblock In {\em Proceedings of the IEEE/CVF Conference on Computer Vision and
  Pattern Recognition}, pages 13528--13537, 2020.

\bibitem{quy2013recognizing}
Trung Quy~Phan, Palaiahnakote Shivakumara, Shangxuan Tian, and Chew Lim~Tan.
\newblock Recognizing text with perspective distortion in natural scenes.
\newblock In {\em Proceedings of the IEEE International Conference on Computer
  Vision}, pages 569--576, 2013.

\bibitem{risnumawan2014robust}
Anhar Risnumawan, Palaiahankote Shivakumara, Chee~Seng Chan, and Chew~Lim Tan.
\newblock A robust arbitrary text detection system for natural scene images.
\newblock {\em Expert Systems with Applications}, 41(18):8027--8048, 2014.

\bibitem{shi2016end}
Baoguang Shi, Xiang Bai, and Cong Yao.
\newblock An end-to-end trainable neural network for image-based sequence
  recognition and its application to scene text recognition.
\newblock {\em IEEE transactions on pattern analysis and machine intelligence},
  39(11):2298--2304, 2016.

\bibitem{shi2018aster}
Baoguang Shi, Mingkun Yang, Xinggang Wang, Pengyuan Lyu, Cong Yao, and Xiang
  Bai.
\newblock Aster: An attentional scene text recognizer with flexible
  rectification.
\newblock {\em IEEE transactions on pattern analysis and machine intelligence},
  41(9):2035--2048, 2018.

\bibitem{srivastava2014dropout}
Nitish Srivastava, Geoffrey Hinton, Alex Krizhevsky, Ilya Sutskever, and Ruslan
  Salakhutdinov.
\newblock Dropout: a simple way to prevent neural networks from overfitting.
\newblock {\em The journal of machine learning research}, 15(1):1929--1958,
  2014.

\bibitem{su2019vl}
Weijie Su, Xizhou Zhu, Yue Cao, Bin Li, Lewei Lu, Furu Wei, and Jifeng Dai.
\newblock Vl-bert: Pre-training of generic visual-linguistic representations.
\newblock In {\em International Conference on Learning Representations}, 2019.

\bibitem{vaswani2017attention}
Ashish Vaswani, Noam Shazeer, Niki Parmar, Jakob Uszkoreit, Llion Jones,
  Aidan~N Gomez, {\L}ukasz Kaiser, and Illia Polosukhin.
\newblock Attention is all you need.
\newblock In {\em Advances in neural information processing systems}, pages
  5998--6008, 2017.

\bibitem{wan2019textscanner}
Zhaoyi Wan, Mingling He, Haoran Chen, Xiang Bai, and Cong Yao.
\newblock Textscanner: Reading characters in order for robust scene text
  recognition.
\newblock {\em arXiv preprint arXiv:1912.12422}, 2019.

\bibitem{wang2011end}
Kai Wang, Boris Babenko, and Serge Belongie.
\newblock End-to-end scene text recognition.
\newblock In {\em 2011 International Conference on Computer Vision}, pages
  1457--1464. IEEE, 2011.

\bibitem{wang2020decoupled}
Tianwei Wang, Yuanzhi Zhu, Lianwen Jin, Canjie Luo, Xiaoxue Chen, Yaqiang Wu,
  Qianying Wang, and Mingxiang Cai.
\newblock Decoupled attention network for text recognition.
\newblock In {\em AAAI}, pages 12216--12224, 2020.

\bibitem{wang2020exploring}
Yizhi Wang and Zhouhui Lian.
\newblock Exploring font-independent features for scene text recognition.
\newblock {\em ECCV}, 2020.

\bibitem{wang2019dsrn}
Yuxin Wang, Hongtao Xie, Zilong Fu, and Yongdong Zhang.
\newblock Dsrn: A deep scale relationship network for scene text detection.
\newblock In {\em IJCAI}, pages 947--953, 2019.

\bibitem{wang2020contournet}
Yuxin Wang, Hongtao Xie, Zheng-Jun Zha, Mengting Xing, Zilong Fu, and Yongdong
  Zhang.
\newblock Contournet: Taking a further step toward accurate arbitrary-shaped
  scene text detection.
\newblock In {\em Proceedings of the IEEE/CVF Conference on Computer Vision and
  Pattern Recognition}, pages 11753--11762, 2020.

\bibitem{xie2019aggregation}
Zecheng Xie, Yaoxiong Huang, Yuanzhi Zhu, Lianwen Jin, Yuliang Liu, and Lele
  Xie.
\newblock Aggregation cross-entropy for sequence recognition.
\newblock In {\em Proceedings of the IEEE Conference on Computer Vision and
  Pattern Recognition}, pages 6538--6547, 2019.

\bibitem{yang2019symmetry}
Mingkun Yang, Yushuo Guan, Minghui Liao, Xin He, Kaigui Bian, Song Bai, Cong
  Yao, and Xiang Bai.
\newblock Symmetry-constrained rectification network for scene text
  recognition.
\newblock In {\em Proceedings of the IEEE International Conference on Computer
  Vision}, pages 9147--9156, 2019.

\bibitem{yin2017scene}
Fei Yin, Yi-Chao Wu, Xu-Yao Zhang, and Cheng-Lin Liu.
\newblock Scene text recognition with sliding convolutional character models.
\newblock {\em arXiv preprint arXiv:1709.01727}, 2017.

\bibitem{yu2020towards}
Deli Yu, Xuan Li, Chengquan Zhang, Tao Liu, Junyu Han, Jingtuo Liu, and Errui
  Ding.
\newblock Towards accurate scene text recognition with semantic reasoning
  networks.
\newblock In {\em Proceedings of the IEEE/CVF Conference on Computer Vision and
  Pattern Recognition}, pages 12113--12122, 2020.

\bibitem{yue2020robustscanner}
Xiaoyu Yue, Zhanghui Kuang, Chenhao Lin, Hongbin Sun, and Wayne Zhang.
\newblock Robustscanner: Dynamically enhancing positional clues for robust text
  recognition.
\newblock {\em eccv}, 2020.

\bibitem{zhan2019esir}
Fangneng Zhan and Shijian Lu.
\newblock Esir: End-to-end scene text recognition via iterative image
  rectification.
\newblock In {\em Proceedings of the IEEE Conference on Computer Vision and
  Pattern Recognition}, pages 2059--2068, 2019.

\bibitem{zhang2020adaptive}
Chuhan Zhang, Ankush Gupta, and Andrew Zisserman.
\newblock Adaptive text recognition through visual matching.
\newblock {\em ECCV}, 2020.

\bibitem{zhou2015icdar}
Xinyu Zhou, Shuchang Zhou, Cong Yao, Zhimin Cao, and Qi Yin.
\newblock Icdar 2015 text reading in the wild competition.
\newblock {\em arXiv preprint arXiv:1506.03184}, 2015.

\end{thebibliography}
}

\end{document}